\documentclass[wcp]{jmlr}

\usepackage{longtable}
\usepackage{indentfirst}
\usepackage{float}

\usepackage{algorithm}
\usepackage{algpseudocode}

\usepackage{booktabs}

\jmlrvolume{XXX}
\jmlryear{2019}
\jmlrworkshop{(under review)}

\title[Data Augmentation Using GANs]{Data Augmentation Using GANs}

  \author{\Name{Fabio Henrique Kiyoiti dos Santos Tanaka} \Email{fabio.henrique.tanaka@usp.br}\\
  \addr University of Sao Paulo
  \AND
  \Name{Claus Aranha} \Email{caranha@cs.tsukuba.ac.jp}\\
  \addr University of Tsukuba, Department of Computer Sciences
 }

\editors{Wee Sun Lee and Taiji Suzuki}

\begin{document}

\maketitle

\begin{abstract}
In this paper we propose the use of Generative Adversarial Networks (GAN) to generate
artificial training data for machine learning tasks. The generation of artificial 
training data can be extremely useful in situations such as imbalanced data sets, 
performing a role similar to SMOTE or ADASYN. It is also useful when the data
contains sensitive information, and it is desirable to avoid using the original data 
set as much as possible (example: medical data). We test our proposal on benchmark 
data sets using different network architectures, and show that a Decision Tree (DT) 
classifier trained using the training data generated by the GAN reached the same, 
(and surprisingly sometimes better), accuracy and recall than a DT trained on the 
original data set.
\end{abstract}
\begin{keywords}
Generative Adversarial Networks, Data Augmentation, Data Imbalance, Privacy
\end{keywords}

\section{Introduction}
When working with machine learning, it is important to have a high-quality data set to train
the algorithm. This means that the data should not only be sufficiently large, to cover as
many cases as possible, but also be a good representation of the reality. Having a good data
set permits the program to have a better model of the underlying characteristics of the data
and makes it easier to generalize these traits. In this scenario, the creation of synthetic
data can be useful for several reasons like oversampling minority classes and generating new
data sets to keep the privacy of the originals.

The first reason for creating synthetic data sets, oversampling the minority class, is
relevant when trying to learn from imbalanced data sets. In many instances, it is common for
databases to have classes that are underrepresented. For example, when dealing with credit
card frauds the ratio between normal
and fraudulent transactions can be 10000 to 1. The same can happen when analysing medical
information where the number of healthy patients is much higher than affected ones. When this
happens, classification algorithms may have difficulties identifying the minority classes
since the program would still have a low error even if it classifies all the minority classes
wrong. To avoid this problem it is possible to augment the minority data though the creation
of new entries by tweaking the original in meaningful ways. This approach not only increases
the representation of the minority class, but it may help to avoid over fitting as well.

The second reason for creating synthetic data sets is to avoid using the original data for
privacy reasons. It is possible that a database contains sensitive information, and working
on it directly could risk it being misused or breached. For example, medical records could
have a lot of personal information about the patients, and even without the names it could be
possible to identify them using a combination of other attributes such as date of birth,
weight, height, etc. Because of this, it is understandable that many regulations exist on the
use of this kind of database, controlling its access. One possible approach to this problem 
is to not use the original data to train the model, but generate a synthetic data set based on
it which sufficiently realistic.

In this research, we studied the use of Generative Adversarial Networks (GAN) to deal with the
two previously mentioned issues. In both cases, we obtained public data sets and generated
synthetic versions of these data sets using different GAN architectures. The quality of 
these synthetic data sets as training data was examined in two ways. First, we compared the 
Accuracy, Precision and Recall obtained by a Decision Tree classifier trained on the 
original data against one trained on the synthetic data. We were surprised to observe that 
in some cases, the classifier trained on the synthetic data achieved better results than the 
one trained in the original data, which suggests that generating synthetic data using GANs
can be a good approach to avoid overfitting. Second, we compared the performance of 
the classifier on imbalanced data sets that were augmented by the GAN, SMOTE~\citep{SMOTE} and
ADASYN~\citep{ADASYN}. In this experiment the GAN improved the results when compared with the
original, imbalanced data set, but did not perform better than SMOTE or ADASYN.

\section{Background}
\subsection{Generating Adversarial Networks (GANs)}
Generating Adversarial Networks, or GANs for short, were first introduced by 
\cite{original_gan}. Since then, many researches have been done using the framework and
Facebook’s AI research director \cite{facebook_gan_recognition} recognized it as "the most
interesting idea in the last 10 years in machine learning" . GANs are a type of generative
model, this means that it can produce new content based on its training data.

GANs can have a variety of applications, developing new molecules for
oncology~\citep{oncology}, and increasing resolution~\citep{increase_resolution}, are some of
them. The most common usage for it, however, is for the generation of new images. Below it is
possible to see an example of faces generated by a GAN based on a dataset composed by photos
of famous people (\cite{famous_people_GAN}).

\begin{figure}[H]
\centering
\includegraphics[width=10cm]{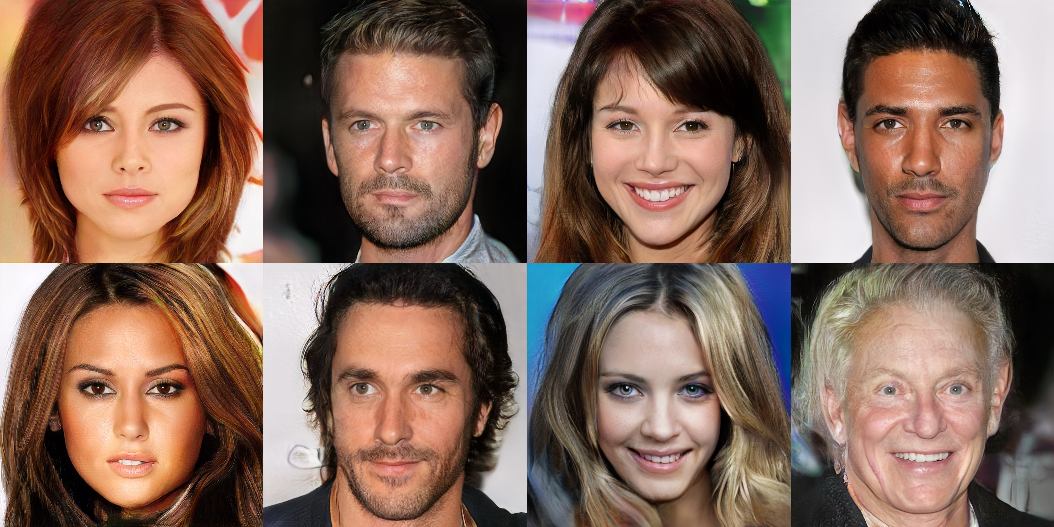}
\caption{\label{fig:gan_example} These people do not exist. Synthetic faces generated by a GAN trained on human pictures~\citep{famous_people_GAN}}
\end{figure}

A GAN is made of 2 Artificial Neural Networks (ANNs) that compete against each other, the
\emph{Generator} and the \emph{Discriminator}. The first creates new data instances while the second
evaluates them for authenticity.

The discriminator is responsible for evaluating the quality of the data created by the
generator. It receives as input data samples from either the original data set, or created by
the generator, and tries to predict the source of the sample. The pseudo-code of its
training is described in Algorithm~\ref{alg:train_discriminator}. 

The generator learns to map a \emph{latent space} to the distribution of the data it aims to reproduce, 
so that when fed with a noise vector from the latent space, it predicts a sample from the estimated 
distribution. The generator is evaluated by the discriminator, meaning that its goal is to create
data samples that are similar to those in the original data set. The pseudo-code for training the 
generator is described in Algorithm~\ref{alg:train_generator}. 

\begin{algorithm}
  \caption{Pseudo-code for training the discriminator}\label{alg:train_discriminator}
  \begin{algorithmic}
  \Require realData\Comment{array of samples from the data set}
  \Require fakeData\Comment{array of samples from the generator}
  \Require discriminator\Comment{discriminator network model}
  \State Set \emph{realDataLabels} as prediction of \emph{realData} from \emph{discriminator};
  \State Set \emph{realLoss} as difference between \emph{realDataLabels} and 1;
  \State Update \emph{discriminator} using \emph{realLoss};
  \State Set \emph{fakeDataLabels} as prediction of \emph{fakeData} from \emph{discriminator};
  \State Set \emph{fakeLoss} as difference between \emph{fakeDataLabels} and 0;
  \State Update \emph{discriminator} using \emph{fakeLoss};
  \end{algorithmic}
\end{algorithm}

\begin{algorithm}
  \caption{Pseudo-code for training the discriminator}\label{alg:train_generator}
  \begin{algorithmic}
  \Require latentVector \Comment{a noise vector sampled from the latent space}
  \Require generator \Comment{generator network model}
  \Require discriminator \Comment{discriminator network model}
  \State Generate \emph{fakeData} as prediction of \emph{latentVector} from \emph{generator};
  \State Set \emph{fakeDataLabels} as prediction of \emph{fakeData} from \emph{discriminator};
  \State Set \emph{fakeLoss} as difference between \emph{fakeDataLabels} and 1;
  \State Update \emph{generator} using \emph{fakeLoss};
  \end{algorithmic}
\end{algorithm}

\begin{figure}[H]
\centering
\includegraphics[width=10cm]{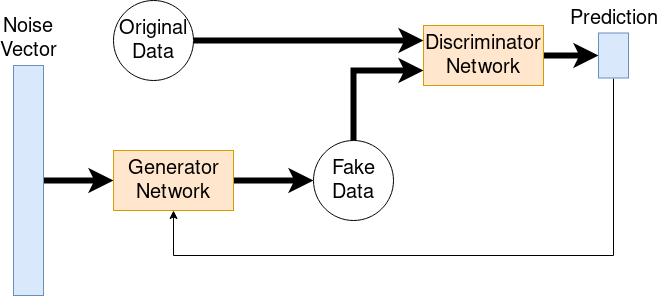}
\caption{\label{fig:GAN working}How a GAN trains both generator and discriminator network at the same time.}
\end{figure}

By training both networks at the same time, they will get better by competing against one another, hence 
the name Generative Adversarial Networks. The discriminator will try to get better at distinguishing 
fake and real data and the generator is going to output data that is progressively closer to the original. 
This iteraction is described in Figure~\ref{fig:GAN working}

The use of GANs have many advantages, they can create high quality data and be modeled to deal with different 
problems. On the other hand, their use have some disadvantages and difficulties as well, some of them are: it 
is hard to generate discrete data (like text), they are hard to train and require large processing power, and 
like any ANN, its model can be unstable or it can overfit the data.

Regarding the two applications addressed in this work (balancing data and generating synthetic data sets),
there is some previous work in the literature using GANs, such as \cite{BAGAN}, \cite{catGAN} and \cite{liver}).
However in all these cases the work was done in images, while we are more interested in standard numerical 
databases (as most of the initial work in GANs was done on images, although this has began to change recently). 
Although the techniques used are similar, there are differences in implementation which are worth
exploring. For example, the use of convolution does not apply to non-image datasets, since the attributes 
of a sample vector do not exhibit positional relationships among themselves.

\subsection{SMOTE and ADASYN}

SMOTE~\citep{SMOTE} and ADASYN~\citep{ADASYN} are two approaches to oversampling the minority 
classes with the goal to balance data sets. In this work, we use their implementation from the 
imbalanced-learn python library~\citep{imblearn}.

Synthetic Minority Over sampling Technique (SMOTE) create synthetic samples based on the position
of the data. First, it randomly selects a point in the minority class, them finds the $k$ nearest 
neighbors of the same class. For each of these pairs a new point is generated in the vector between 
them, this new point is located in a random percent of the way from the original point.

\begin{figure}[H]
\centering
\includegraphics[width=10cm]{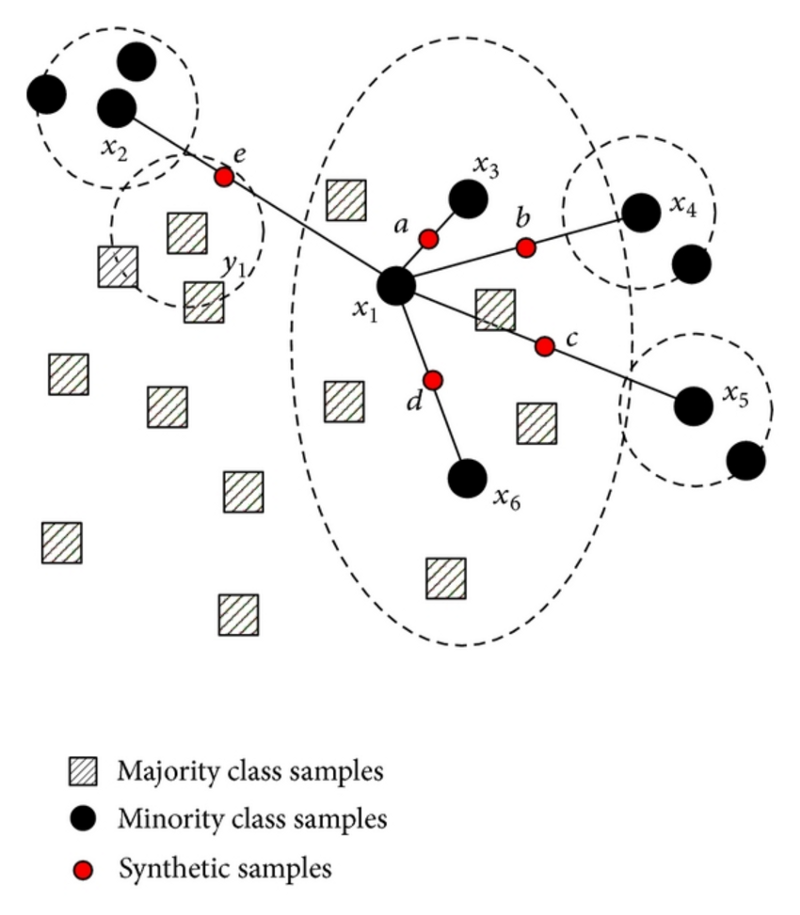}
\caption{\label{fig:smote}Example of SMOTE (from \cite{SMOTE_image})}
\end{figure}

ADASYN is similar to SMOTE, and is derived from it. They function on the same way but, after creating 
the samples, ADASYN adds a random small bias to the points, making them not linearly correlated to 
their parents. Even though this is a small change it increases the variance in the synthetic data.

\section{Proposal and Experimental Design}

We evaluate the utility of GANs for generating synthetic numerical data sets in two different domains:
To train a classifier using purely synthetic data, and to balance a data set by oversampling the 
minority class using synthetic data.

For the first domain, we will compare the performance of the classifier on the original data set and 
the data sets created by variations in our GAN architecture. The idea is that if we can obtain a synthetic
data set that is similar enough to the original data set, we can completely avoid training the classifier on 
the original data set. Yes, training the GAN itself is still necessary. However, many different classifiers 
for different tasks might need to be trained by different entities, and using synthetic data will 
still reduce the need to distribute the original datain this case.

For the second domain, we will compare the performance of the classifier on imbalanced data oversampled 
with GAN, SMOTE and ADASYN, as well as non-oversampled data as a baseline. While SMOTE and ADASYN both 
produce good results, they tend to not generalize well sparse data and outliers. We wonder if 
oversampling using GAN can overcome these issues.

For reproducibility purposes, the code of our GAN architecture is available at our github 
repository~\footnote{ \url{https://github.com/fhtanaka/directed_research_CS_2018}}.

\subsection{GAN architecture and parameter Choice}

In both experimental domains, we use the same general GAN architecture to generate the synthetic data.
Our GAN architecture has the following parameters:

\begin{itemize}
    \setlength\itemsep{-0.5em}
    \item \textit{leaky ReLU} as the activation function with a negative slope of 0.2.
    \item batch size = 5
    \item learning rate = 0.0002
    \item Use of dropout in the generator with a probability of 0.3.
    \item Binary cross-entropy as the loss function.
    \item Adam as the optimizer algorithm.
    \item No convolution layers.
    \item In the generator, if there are more than one layer, they are ordered in ascending size.
    \item In the discriminator, if there are more than one layer, they are ordered in descending size.
\end{itemize}

Leaky ReLU, Adam optimizer and the use of dropout were chosen because they are the standard for this kind of 
problem~\citep{GAN_tips}. There are no convolutional layers because the input is not an image. Binary 
cross-entropy loss is used because it is the most fit to measure the performance of a model whose output is a 
probability between 0 and 1. 

We tested six different configurations varying the number of layers and the number of nodes in each layer, 
and all the results are included in the following sections. In this experiment, we intentionally use a 
simple network and configuration, to focus on the basic proposal of generating synthetic numerical databases.

\subsection{Classifier}

For both domains of this experiment, we use a Decision Tree classifier as the classifier to test the 
quality of the synthetic training data generated. We use Decision Tree classifier implementation from the 
\emph{sklearn} library. This choice was made because Decision Trees are simple to understand and
interpret since they can be visualized, they also require little to no data preparation.

\subsection{Databases}
The experiments on this research were done using the following 3 different data sets:

\begin{itemize}
    \item \textbf{Pima Indians Diabetes Database (\cite{diabetes_db}):} This data set consists of 8 independent 
    variables and one target dependent class that represents if the patient has Diabetes or not. It is composed by 
    768 samples, in which 268 patients present the disease.
    
    \item \textbf{Breast Cancer Wisconsin (Diagnostic) Data Set (\cite{cancer_db}):} The features of this data set 
    are computed from a digitized image of a fine needle aspirate (FNA) of a breast mass. They describe characteristics 
    of the cell nuclei present in the image. The data set has a target class to determine if the cancer is benign or 
    malign. There are 357 benign and 212 malignant samples.
    
    \item \textbf{Credit Card Fraud Detection (\cite{creditcard_db}):} This data set contains transactions made by 
    credit cards in September 2013 by European cardholders. It presents transactions that occurred in two days, 
    where there are 492 frauds out of 284,807 transactions. The data set is highly imbalanced, the positive class 
    (frauds) account for only 0.172\% of all transactions.
    
\end{itemize}

The goal class on the \textit{Breast Cancer Wisconsin (Diagnostic) Data Set} was represented by a char, 
"B" = benign and "M" = malignant but this label was changed to 0 and 1 respectively to be consistent with the other 
examples. With the exception of this case, all attributes are fully numeric, which was the reason for this choice 
of data sets. Working with fully numeric examples allows the GAN to generate discrete results. 
They are also faster to compute and train when compared to image data sats. Each of these data sets was 
divided in two subsets, one for training the GAN (the first 70\% of the original data) and other for 
testing it (the remaining 30\%). A summary of the data sets is in table~\ref{tab:dbs}.

\begin{table}[H]
    \centering
    \resizebox{\textwidth}{!}  
    {%
    \begin{tabular}{||c c c c||}
        \hline
            Database Name & Number of features & Size & Label Distribution\\
        \hline\hline
            Pima Indians Diabetes Database & 9 & 768 & No diabetes: 500, Diabetes: 268 \\
            Breast Cancer Wisconsin (Diagnostic) Data Set & 32 & 569 & Benign: 357, Malignant: 212\\
            Credit Card Fraud Detection & 31 & 284807 & Non-frauds: 284315, Frauds: 492 \\
        \hline
    \end{tabular}
    }
    \caption{Databases used in this research}
    \label{tab:dbs}
\end{table}

Notice that, before using these databases, their attribute values were all scaled to be in the
interval [0,1] by the min-max method. This was done because it makes the range for all attributes 
to be the same, preventing one of them to dominate the others because of its scale. This reduces the range 
of values that the generator has to produce as well.

\section{Results and Analysis}

In the results below, the synthetic databases generated by the 6 variations of the GAN architecture 
are described using the names shown in table~\ref{tab:dbs_names}.

\begin{table}[H]
    \centering
    \begin{tabular}{||l l||}
        \hline
            Data set name & Architecture of the GAN\\
        \hline\hline
            original data       & The first 70\% of the original database \\
            256/512/1024   & Generated by a GAN with 3 hidden layers with size 256, 512 and 1024 \\
            256/512   & Generated by a GAN with 2 hidden layers with size 256 and 512\\
            256  & Generated by a GAN with 1 hidden layer with size 256\\
            128/256/512 & Generated by a GAN with 3 hidden layers with size 128, 256 and 512 \\
            128/256 & Generated by a GAN with 2 hidden layers with size 128 and 256\\
            128 & Generated by a GAN with 1 hidden layer with size 128\\
        \hline
    \end{tabular}
    \caption{Names used for the synthetic and original data sets in the results}
    \label{tab:dbs_names}
\end{table}

\subsection{Experiment 1: Training the classifier using fully synthetic data}
To evaluate the creation of synthetic data the experiments were done using the following steps:

\begin{enumerate}
    \setlength\itemsep{-0.5em}
    \item Trained the GAN using the full training subset of the 
    original database for 1500 epochs.
    \item Used the newly trained GAN to generate a new 
    synthetic data set with the exact size of the original.
    \item Since the GANs generated the classification label as 
    a continuous value between 0 and 1, this value has to be turned to a discrete by rounding it to the nearest integer.
    \item The new data set is used to train a classification tree.
    \item The tree is tested using the test subset of the 
    original data set.
\end{enumerate}

It is important to note that the GAN was trained using the label classes as well. This means that the data generated 
can have any of the classes and the GAN itself defines how each data point should be classified.

\begin{figure}[H]
\centering
\includegraphics[width=\textwidth]{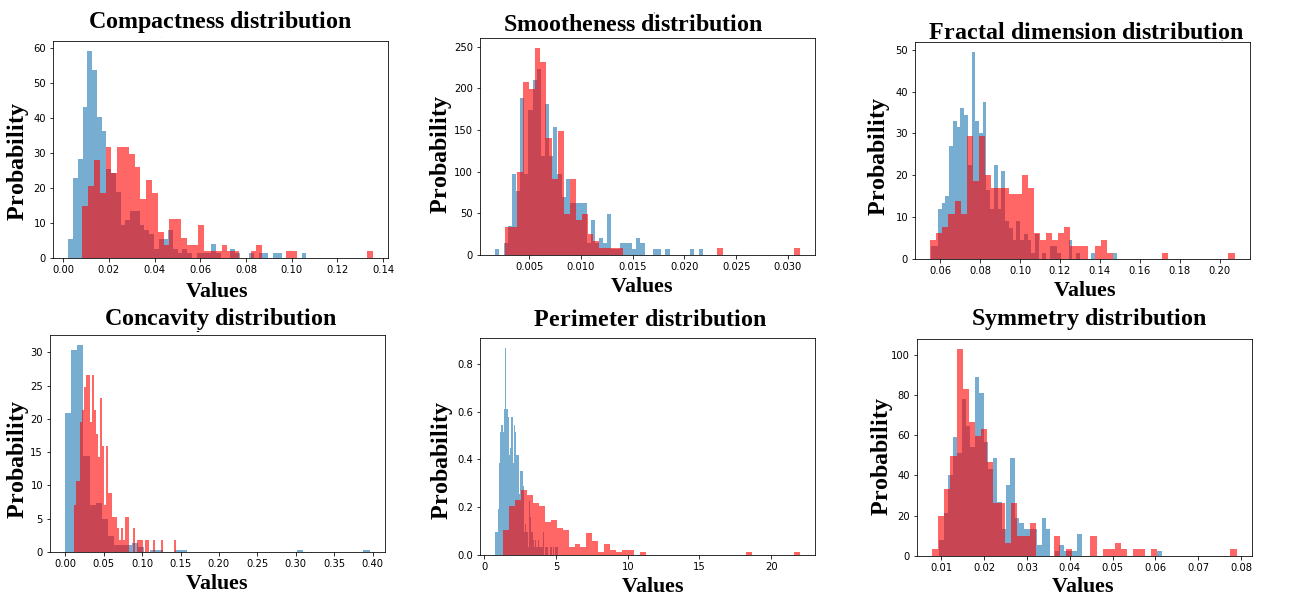}
\caption{\label{fig:data_distribution}Comparison between the distribution of some attributes in the original (blue) and synthetic (red) cancer data set}
\end{figure}

The tests were done using the diabetes and cancer databases. Both are not very unbalanced and have less than 
1000 entries. Figure~\ref{fig:data_distribution} shows the distribution of some attributes of the new synthetic cancer 
data set compared with the original.

Tables~\ref{tab:breast_cancer_early_results} and~\ref{tab:diabetes_early_results} show an early result 
where a single classification test is done using synthetic data generated after 1500 epochs of training.
Tables~\ref{tab:breast_cancer_test_results} and~\ref{tab:data_tests results} are a more in-depth analysis 
of the results. We generate 20 different synthetic data sets, and perform the previous classification experiment 
on each of them. The values between parenthesis are the standard deviations of the results.

Tables~\ref{tab:breast_cancer_test_distance} and~\ref{tab:diabetes_test_distance} show the mean Euclidean distance 
between the points in the the synthetic database and the closest point to it in the original. This is used to 
evaluate the similarity between the two data sets. Again, the average results of 20 data sets is displayed.

\begin{table}[H]
    \centering
    \begin{tabular}{||l c c||}
        \hline
            Database & Label Proportion & Test Accuracy\\
        \hline\hline
            original data       & 56.53/43.47 & 0.888 \\
            256/512/1024   & 52.26/47.74 & 0.818 \\
            256/512   & 56.28/43.72 & 0.941 \\
            256   & 56.78/43.22 & 0.906 \\
            128/256/512 & 54.02/45.98 & 0.953 \\
            128/256 & 58.04/41.96 & 0.935 \\
            128 & 54.27/45.73 & 0.912 \\
        \hline
    \end{tabular}
    \caption{Label proportion and accuracy of the tests in the Cancer database}
    \label{tab:breast_cancer_early_results}
\end{table}

\begin{table}[ht]
    \centering
    \begin{tabular}{||l c c||}
        \hline
            Database & Label Proportion & Test Accuracy\\
        \hline\hline
            original data       & 64.8/35.2     & 0.748 \\
            256/512/1024   & 71.69/28.31   & 0.7 \\   
            256/512   & 67.23/32.77   & 0.548 \\
            256   & 67.6/32.4     & 0.748 \\
            128/256/512 & 60.15/39.85   & 0.661 \\
            128/256 & 65.18/34.82   & 0.739 \\
            128 & 68.9/31.1     & 0.697 \\ 
        \hline
    \end{tabular}
    \caption{Label proportion and accuracy of the tests in the Diabetes database}
    \label{tab:diabetes_early_results}
\end{table}

\begin{table}[H]
    \centering
    \resizebox{\textwidth}{!} {%
    \begin{tabular}{||l c c c||}
        \hline
            Database & Accuracy & Precision & Recall\\
        \hline\hline
            Original data & 0.888 & 0.679 & 0.974 \\
            256/512/1024 & 0.91 (0.042) & 0.786 (0.127) & 0.888 (0.054) \\
            256/512 & {\bf 0.935} (0.03) & {\bf 0.853} (0.107) & {\bf 0.896} (0.086) \\
            256 & 0.907 (0.053) & 0.772 (0.126) & 0.904 (0.071) \\
            128/256/512 & 0.869 (0.066) & 0.702 (0.144) & 0.821 (0.148) \\
            128/256 & 0.896 (0.048) & 0.74 (0.122) & 0.908 (0.086) \\
            128 & 0.906 (0.054) & 0.775 (0.131) & 0.894 (0.055) \\
        \hline
    \end{tabular}%
    }
    \caption{Mean and standard deviation (n=20) of the classification results in the cancer database.}
    \label{tab:breast_cancer_test_results}
\end{table}

\begin{table}[H]
    \centering
    \resizebox{\textwidth}{!} {%
    \begin{tabular}{||l c c c||}
        \hline
            Database & Accuracy & Precision & Recall\\
        \hline\hline
            Original data & {\bf 0.748} & {\bf 0.784} & 0.367 \\
            256/512/1024 & 0.682 (0.064) & 0.545 (0.093) & 0.534 (0.206) \\
            256/512 & {\bf 0.706} (0.05) & {\bf 0.582} (0.078) & {\bf 0.584} (0.097) \\
            256 & 0.601 (0.097) & 0.438 (0.118) & 0.438 (0.213) \\
            128/256/512 & 0.685 (0.058) & 0.568 (0.109) & 0.544 (0.158) \\
            128/256 & 0.639 (0.094) & 0.507 (0.106) & 0.579 (0.185) \\
            128 & 0.653 (0.086) & 0.51 (0.117) & 0.462 (0.219) \\
        \hline
    \end{tabular}%
    }
    \caption{Mean and standard deviation (n=20) of the classification results in the diabetes database.}
    \label{tab:data_tests results}
\end{table}

\begin{table}[H]
    \centering
    \begin{tabular}{||l c||}
        \hline
            Database & Distance\\
        \hline\hline
            256/512/1024 & 0.345 (0.156) \\
            256/512 & 0.314 (0.121) \\
            256 & 0.328 (0.135) \\
            128/256/512 & 0.371 (0.190) \\
            128/256 & 0.331 (0.128) \\
            128 & 0.342 (0.143) \\
        \hline
    \end{tabular}
    \caption{Mean and standard deviation of the euclidean distance between the points in the synthetic and original cancer database.}
    \label{tab:breast_cancer_test_distance}
\end{table}

\begin{table}[H]
    \centering
    \begin{tabular}{||l c||}
        \hline
            Database & Distance\\
        \hline\hline
            256/512/1024 & 0.158 (0.071) \\
            256/512 & 0.17 (0.071) \\
            256 & 0.208 (0.106) \\
            128/256/512 & 0.172 (0.069) \\
            128/256 & 0.187 (0.080) \\
            128 & 0.189 (0.088) \\
        \hline
    \end{tabular}
    \caption{Mean and standard deviation of the euclidean distance between the points in the synthetic and original diabetes database.}
    \label{tab:diabetes_test_distance}
\end{table}

\subsection{Analysis of the results for the fully synthetic data set}

The early results in both the cancer and the diabetes data sets showed that GANs can produce synthetic data 
with a similar class and attribute distribution as the original data, without an explicit instruction to do so.
This is the result of training the GAN with the complete training set, without separation of the classes. 
This indicates that for practical use it is not necessary to separate the data in classes before training, 
reducing the burden for the data provider, and the need of domain knowledge for the data user.

Considering the mean accuracy, precision and recall results, the 256/512 layer architecture had the best overall
outcomes in both data sets, with statistically significant difference ($p < 0.05$) for the accuracy measure.
It is hard to explain precisely why this architecture was better, but we recommend it's use until further experiments 
provide more information. The following considerations will use this architecture as the baseline.

When we compare the accuracy obtained by training the classifier with the synthetic data with the 
accuracy obtained by training the classifier with the original data, the accuracy of the synthetic
data is the close to the original data in the diabetes dataset, and better than the original data
in the cancer data set. This is a good indicator that using synthetic GAN data as training data for 
classifiers is a promising technique. 

However, when classifying medial data, we are more interested in the Recall than the Accuracy, since 
false negatives can be disastrous when deciding if a patient needs a treatment or has a certain diagnostic.
In this aspect, the synthetic data from GAN obtained great results in the diabetes data set, while the 
results in the cancer data set were slightly lower than the recall on the original data set.

We also want to know if synthetic GAN data can be used to improve the privacy of handling medical 
data sets in training models. Since the experiments were done in a dataset where all attributes are scaled 
to the 0,1 interval, the euclidean distance between the data points in the original and synthetic data set
are a good estimation of similarity between the data sets. Tables~\ref{tab:breast_cancer_test_distance}
and~\ref{tab:diabetes_test_distance} show this distance. The large distance observed in for each 
model, specially for the cancer database, indicates that it is unlikely that an ill-intentioned user with 
access to the synthetic data set would be able to use it to pinpoint exact information about any one 
user in the original data set.

As a summary, the classifier trained with synthetic data from the GAN was able to reproduce the results 
obtained from training on the original data fairly well. On the other hand, depending on the characteristics
of the problems, this approach could result in an increase of false negatives or false positives. More 
effort should be spent to try to understand exactly when these differences take place, and what attributes 
of the problem make the GANs have better or worse precision and recall.

\subsection{Experiment 2: Oversampling the Minority Class with Synthetic Data}

To evaluate the effects of balancing a data set by oversampling the minority class using synthetic 
data produced from a GAN, our experiment uses the following steps:

\begin{enumerate}
    \setlength\itemsep{-0.5em}
    \item Separated the training set based on the target class (For example, the credit card
    database was separated in non-frauds and frauds).
    \item Trained the GAN using only the minority class data. The label was included as an attribute in the training.
    \item Used the GAN to add new entries to the training data set until it becomes balanced.
    \item Used the newly balanced training data set to train the classifier.
    \item The classifier was tested on two databases: the original test set and a balanced version of it obtained by undersampling the majority class.
\end{enumerate}

Tables~\ref{tab:unbalanced_test} and~\ref{tab:balanced_test} show the mean results from 5 repetitions,
with the standard deviation in parenthesis. The smaller number of repetitions is due to the large 
quantity of data in the data sets. As the SMOTE and ADASYM implementation in the imblearn library~\citep{imblearn} 
always produced the same output, the standard deviation for these two cases is zero.

\begin{table}[H]
    \centering
    \resizebox{\textwidth}{!} {%
    \begin{tabular}{||l c c c||}
        \hline
            Database & Accuracy & Precision & Recall\\
        \hline\hline
            Original & {\bf 0.999} & {\bf 0.896} & 0.556 \\
            SMOTE & 0.958 & 0.026 & {\bf 0.861} \\
            ADASYN & 0.958 & 0.026 & {\bf 0.861} \\
            128 & 0.798 (0.372) & 0.051 (0.029) & 0.806 (0.042) \\
            256 & {\bf 0.986} (0.005) & {\bf 0.077} (0.031) & 0.789 (0.018) \\
            128/256 & 0.974 (0.01) & 0.045 (0.02) & 0.82 (0.028) \\
            256/512 & 0.964 (0.017) & 0.033 (0.013) & 0.808 (0.069) \\
        \hline
    \end{tabular}%
    }
    \caption{Classification results from testing on the \textbf{imbalanced} test set}
    \label{tab:unbalanced_test}
\end{table}

\begin{table}[H]
    \centering
    \resizebox{\textwidth}{!} {%
    \begin{tabular}{||l c c c||}
        \hline
            Database & Accuracy & Precision & Recall\\
        \hline\hline
            Original & 0.782 & {\bf 1.0} & 0.565 \\
            SMOTE & 0.912 & 0.959 & {\bf 0.861} \\
            ADASYN & {\bf 0.921} & 0.979 & {\bf 0.861} \\
            128 & 0.807 (0.165) & 0.89 (0.202) & 0.806 (0.042) \\
            256 & 0.894 (0.01) & {\bf 0.998} (0.005) & 0.789 (0.018) \\

            128/256 & 0.902 (0.012) & 0.981 (0.015) & 0.82 (0.028) \\
            256/512 & 0.888 (0.032) & 0.962 (0.018) & 0.808 (0.069) \\
        \hline
    \end{tabular}%
    }
    \caption{Classification results from testing on the \textbf{balanced} test set}
    \label{tab:balanced_test}
\end{table}

\subsection{Analysis of the "oversampling on the minority class" experiment}

First we that the GAN with only one layer of 128 nodes displayed
particularly bad results when compared to the other GAN variations. Clearly there 
was a catastrophic divergence in the training of this GAN and we should strive to be able
to identify and correct these cases.

For all the methods, oversampling the minority class improved the recall score and worsened
the precision score. This is to be expected, because the original data set is so imbalanced that
the tree trained on it predicts almost all samples as in the negative (non-fraud) class.
Also, we observe that for all methods, the balanced test set shows better results than the 
imbalanced test set. Again, this makes sense because the classifier has an easier time to identify 
positive labels. That said, more importance should be given to the imbalanced test set 
results, as we cannot expect that the test data in actual applications will be balanced. 

With that in mind, when dealing with the unbalanced test set, the GAN had better 
accuracy and precision and worse recall when compared to ADASYN and SMOTE. The importance of
each of these results may vary depending on the problem domain but, in the case of fraud evaluation on 
credit cards, it is much more important to identify frauds. As such, a good recall is 
desired. In this aspect ADASYN and SMOTE should be prefferred to our GAN framework for data synthesis.

Even so, using GAN for data augmentation improved the results when compared to just using the 
original data set. With a little more work on the GAN architecture and parameters, it is not unreasonable
to expect comparable results to the other two algorithms.

\section{Conclusions and Future Work}

In this work we proposed the use of a GAN to generate synthetic data for a numeric dataset, with the objective 
to train a classifier without using the original data set, and to oversample the minority class in a 
imbalanced classification scenario. In both scenarios the data sets generated by GAN were suitable 
for the tasks proposed. 

Training the classifier using only GAN synthetic data in the balanced scenario showed better accuracy and 
precision than training on the original data set. On the imbalanced scenario, the GAN synthetic data 
performed better than the original data, but did not perform better than SMOTE or ADASYN. One benefit of 
using GANs is that the user does not have to define any rules or constraints in general. 

As an initial inquiry, this research limits itself to one GAN architecture and 6 variations of network
depth and width. We also used only three data sets that are widely used as classification benchmarks. 
Our future work will analyse many different network architectures as well as data sets with different 
characteristics. We are specifically interested in investigating the use of an autoencoder to initialize 
the GAN network, as suggested in BAGAN by~/cite{BAGAN}, and in investigating the influence of the 
number of classes and attributes in the data set and the final results. Other aspects that can 
influence the results are the quality and distribution of samples in the data set, and the influence 
of mislabeled data and missing data.


\bibliography{ACML_GAN}

\begin{thebibliography}{16}
\providecommand{\natexlab}[1]{#1}
\providecommand{\url}[1]{\texttt{#1}}
\expandafter\ifx\csname urlstyle\endcsname\relax
  \providecommand{\doi}[1]{doi: #1}\else
  \providecommand{\doi}{doi: \begingroup \urlstyle{rm}\Url}\fi

\bibitem[Chawla et~al.(2011)Chawla, Bowyer, Hall, and Kegelmeyer]{SMOTE}
N.~V. Chawla, K.~W. Bowyer, L.~O. Hall, and W.~P. Kegelmeyer.
\newblock Smote: Synthetic minority over-sampling technique.
\newblock \emph{Journal Of Artificial Intelligence Research, Volume 16, pages
  321-357, 2002}, 2011.
\newblock \doi{10.1613/jair.953}.

\bibitem[Chintala et~al.(2016)Chintala, Denton, Arjovsky, and
  Mathieu]{GAN_tips}
Soumith Chintala, Emily Denton, Martin Arjovsky, and Michael Mathieu.
\newblock How to train a gan? tips and tricks to make gans work, 2016.
\newblock URL \url{https://github.com/soumith/ganhacks}.

\bibitem[Dal~Pozzolo et~al.(2017)Dal~Pozzolo, Boracchi, Caelen, Alippi, and
  Bontempi]{creditcard_db}
Andrea Dal~Pozzolo, Giacomo Boracchi, Olivier Caelen, Cesare Alippi, and
  Gianluca Bontempi.
\newblock Credit card fraud detection: A realistic modeling and a novel
  learning strategy.
\newblock \emph{IEEE Transactions on Neural Networks and Learning Systems},
  PP:\penalty0 1--14, 09 2017.
\newblock \doi{10.1109/TNNLS.2017.2736643}.

\bibitem[Dua and Graff(2017)]{cancer_db}
Dheeru Dua and Casey Graff.
\newblock {UCI} machine learning repository, 2017.
\newblock URL \url{http://archive.ics.uci.edu/ml}.

\bibitem[Frid-Adar et~al.(2018)Frid-Adar, Klang, Amitai, Goldberger, and
  Greenspan]{liver}
Maayan Frid-Adar, Eyal Klang, Michal Amitai, Jacob Goldberger, and Hayit
  Greenspan.
\newblock Synthetic data augmentation using gan for improved liver lesion
  classification.
\newblock In \emph{2018 IEEE 15th International Symposium on Biomedical Imaging
  (ISBI 2018)}, pages 289--293. IEEE, 2018.

\bibitem[Goodfellow et~al.(2014)Goodfellow, Pouget-Abadie, Mirza, Xu,
  Warde-Farley, Ozair, Courville, and Bengio]{original_gan}
Ian Goodfellow, Jean Pouget-Abadie, Mehdi Mirza, Bing Xu, David Warde-Farley,
  Sherjil Ozair, Aaron Courville, and Yoshua Bengio.
\newblock Generative adversarial nets.
\newblock In \emph{Advances in neural information processing systems}, pages
  2672--2680, 2014.

\bibitem[He et~al.(2008)He, Bai, Garcia, and Li]{ADASYN}
Haibo He, Yang Bai, Edwardo~A. Garcia, and Shutao Li.
\newblock Adasyn: Adaptive synthetic sampling approach for imbalanced learning.
\newblock \emph{2008 IEEE International Joint Conference on Neural Networks
  (IEEE World Congress on Computational Intelligence)}, pages 1322--1328, 2008.

\bibitem[Hu and Li(2013)]{SMOTE_image}
Feng Hu and Hang Li.
\newblock A novel boundary oversampling algorithm based on neighborhood rough
  set model: Nrsboundary-smote.
\newblock \emph{Mathematical Problems in Engineering}, 2013, 11 2013.
\newblock \doi{10.1155/2013/694809}.

\bibitem[Kadurin et~al.(2017)Kadurin, Aliper, Kazennov, Mamoshina, Vanhaelen,
  Khrabrov, and Zhavoronkov]{oncology}
Artur Kadurin, Alexander Aliper, Andrey Kazennov, Polina Mamoshina, Quentin
  Vanhaelen, Kuzma Khrabrov, and Alex Zhavoronkov.
\newblock The cornucopia of meaningful leads: Applying deep adversarial
  autoencoders for new molecule development in oncology.
\newblock \emph{Oncotarget}, 8\penalty0 (7):\penalty0 10883, 2017.

\bibitem[Karras et~al.(2018)Karras, Aila, Laine, and
  Lehtinen]{famous_people_GAN}
Tero Karras, Timo Aila, Samuli Laine, and Jaakko Lehtinen.
\newblock Progressive growing of {GAN}s for improved quality, stability, and
  variation.
\newblock In \emph{International Conference on Learning Representations}, 2018.
\newblock URL \url{https://openreview.net/forum?id=Hk99zCeAb}.

\bibitem[LeCun(2017)]{facebook_gan_recognition}
Yann LeCun.
\newblock What are some recent and potentially upcoming breakthroughs in deep
  learning?, July 2017.
\newblock URL
  \url{www.quora.com/What-are-some-recent-and-potentially-upcoming-breakthroughs-in-deep-learning}.

\bibitem[Ledig et~al.(2017)Ledig, Theis, Husz{\'a}r, Caballero, Cunningham,
  Acosta, Aitken, Tejani, Totz, Wang, et~al.]{increase_resolution}
Christian Ledig, Lucas Theis, Ferenc Husz{\'a}r, Jose Caballero, Andrew
  Cunningham, Alejandro Acosta, Andrew Aitken, Alykhan Tejani, Johannes Totz,
  Zehan Wang, et~al.
\newblock Photo-realistic single image super-resolution using a generative
  adversarial network.
\newblock In \emph{Proceedings of the IEEE conference on computer vision and
  pattern recognition}, pages 4681--4690, 2017.

\bibitem[Lema{{\^i}}tre et~al.(2017)Lema{{\^i}}tre, Nogueira, and
  Aridas]{imblearn}
Guillaume Lema{{\^i}}tre, Fernando Nogueira, and Christos~K. Aridas.
\newblock Imbalanced-learn: A python toolbox to tackle the curse of imbalanced
  datasets in machine learning.
\newblock \emph{Journal of Machine Learning Research}, 18\penalty0
  (17):\penalty0 1--5, 2017.
\newblock URL \url{http://jmlr.org/papers/v18/16-365.html}.

\bibitem[Mariani et~al.(2018)Mariani, Scheidegger, Istrate, Bekas, and
  Malossi]{BAGAN}
Giovanni Mariani, Florian Scheidegger, Roxana Istrate, Costas Bekas, and
  Cristiano Malossi.
\newblock Bagan: Data augmentation with balancing gan, 2018.

\bibitem[Smith et~al.(1988)Smith, Everhart, Dickson, W.C.Knowler, and
  Johannes]{diabetes_db}
J.W. Smith, J.E. Everhart, W.C. Dickson, W.C.Knowler, and R.S. Johannes.
\newblock Using the adap learning algorithm to forecast the onset of diabetes
  mellitus.
\newblock In \emph{Proceedings of the Symposium on Computer Applications and
  Medical Care}, pages 261--265, 1988.
\newblock URL
  \url{https://www.kaggle.com/uciml/pima-indians-diabetes-database}.

\bibitem[Springenberg(2015)]{catGAN}
Jost~Tobias Springenberg.
\newblock Unsupervised and semi-supervised learning with categorical generative
  adversarial networks.
\newblock \emph{arXiv preprint arXiv:1511.06390}, 2015.

\end{thebibliography}

\end{document}